\documentclass[manuscript]{acmart} 
\AtBeginDocument{%
  }

\setcopyright{acmlicensed}
\copyrightyear{2025}
\acmYear{2025}
\acmDOI{XXXXXXX.XXXXXXX}

\acmJournal{POMACS}
\acmVolume{37}
\acmNumber{4}
\acmArticle{111}
\acmMonth{8}
\usepackage{booktabs}
\usepackage{float}
\usepackage{multirow}
\usepackage{tabularx}   
\usepackage{caption}
\usepackage{makecell}
\usepackage{CJKutf8}
\newcommand{\varcell}[1]{\textcolor{black}{#1}}
\newcommand{\varbest}[1]{\textbf{\textcolor{black}{#1}}}
\newcolumntype{B}{>{\color{black}}c}

\begin{document}

\title{Bridging Linguistic Gaps: Cross-Lingual Mapping in Pre-Training and Dataset for Enhanced Multilingual LLM Performance}

\author{WEIHUA ZHENG}
\email{weihua_zheng@mymail.sutd.edu.sg}
\orcid{0009-0009-3855-2998}
\affiliation{%
  \institution{Singapore University of Technology and Design}
  \country{Singapore}
}

\author{CHANG LIU}
\affiliation{%
  \institution{ByteDance}
  \country{Singapore}}
\email{Liuc0058@e.ntu.edu.sg}

\author{ZHENGYUAN LIU}
\affiliation{%
  \institution{Agency for Science, Technology and Research}
  \country{Singapore}
}
\email{liu_zhengyuan@i2r.a-star.edu.sg}

\author{XIN HUANG}
\affiliation{%
 \institution{Agency for Science, Technology and Research}
 \country{Singapore}}

\author{KUI WU}
\affiliation{%
  \institution{Agency for Science, Technology and Research}
  \country{Singapore}}
\email{wuk@i2r.a-star.edu.sg}

\author{MUHAMMAD HUZAIFAH MD SHAHRIN}
\affiliation{%
  \institution{Agency for Science, Technology and Research}
  \country{Singapore}}
\email{huzaifah_md_shahrin@i2r.a-star.edu.sg}

\author{AITI AW}
\affiliation{%
  \institution{Agency for Science, Technology and Research}
  \country{Singapore}}
\email{aaiti@i2r.a-star.edu.sg}

\author{ROY KA-WEI LEE}
\affiliation{%
  \institution{Singapore University of Technology and Design}
  \country{Singapore}}
\email{s.roylee@gmail.com}

\renewcommand{\shortauthors}{Zheng et al.}

\begin{abstract}
Multilingual Large Language Models (LLMs) struggle with cross-lingual tasks due to data imbalances between high-resource and low-resource languages and the monolingual bias in pre-training. Existing methods, such as bilingual fine-tuning and contrastive alignment, improve cross-lingual performance but often require extensive parallel data or suffer from instability. To address these challenges, we introduce a Cross-Lingual Mapping Task in the pre-training phase, which enhances cross-lingual alignment without compromising monolingual fluency. Our approach bi-directionally maps languages within the LLM's embedding space, improving both language generation and comprehension. We further introduce a Language Alignment Coefficient to robustly quantify cross-lingual consistency, even in limited-data scenarios. Experimental results on machine translation (MT), cross-lingual natural language understanding (CLNLU), and cross-lingual question answering (CLQA) show that our model achieves up to 11.9 BLEU score gains in MT, an increase of 6.72 in CLQA BERTScore-Precision and more than a 5\% increase in CLNLU accuracy over strong multilingual baselines. Our findings highlight the potential of embedding cross-lingual objectives into pre-training, improving multilingual LLMs.
\end{abstract}

\ccsdesc[500]{Computing methodologies~Natural language generation}

\keywords{Cross-Lingual, Large Language Models, Low-resource Language}


\maketitle

\section{Introduction}

\paragraph{Motivation.} Recent advancements in Large Language Models (LLMs) have significantly improved Natural Language Processing (NLP) capabilities, achieving state-of-the-art results in tasks such as cross-lingual question answering (CLQA), text summarization, and machine translation (MT)~\cite{devlin2018bert}. Multilingual LLMs, including mBERT~\cite{devlin2018bert}, mT5~\cite{xue2020mt5}, and Llama3~\cite{dubey2024llama3herdmodels}, have extended these advancements to multilingual settings. Nevertheless, these models remain suboptimal in cross-lingual tasks requiring text generation and comprehension~\cite{zeng2023glm130bopenbilingualpretrained}, particularly in MT and cross-lingual summarization (CLSum), where maintaining semantic fidelity and linguistic coherence remains challenging. A key factor behind this limitation is the imbalance in training data, with high-resource languages dominating pre-training corpora~\cite{gao2024multilingualpretraininginstructiontuning}. Consequently, low-resource languages are underrepresented, leading to weaker cross-lingual generalization. Moreover, standard pre-training predominantly relies on monolingual next-token prediction, reinforcing fluency in individual languages but limiting cross-lingual transfer.

Several approaches have attempted to address these limitations. Continued pre-training on low-resource languages improves cross-lingual proficiency~\cite{xu2024paradigmshiftmachinetranslation}, while word-level substitution and alignment techniques enhance language transfer~\cite{NEURIPS2019_c04c19c2, tang2022alignmlmwordembeddingalignment, wu-dredze-2020-explicit}. However, these methods struggle with polysemy, grammatical inconsistencies, and code-switching. Furthermore, techniques like bilingual sentence masking, though promising, are often incompatible with decoder-only LLMs that generate text autoregressively. These challenges highlight the need for a more effective pre-training paradigm that explicitly integrates cross-lingual alignment while preserving monolingual fluency.

\paragraph{Research Objectives.} This study aims to address these research gaps to improve the cross-lingual capabilities of multilingual LLMs. Specifically, we seek to (i) develop an effective pre-training strategy that enhances cross-lingual alignment while preserving monolingual fluency, (ii) introduce a robust evaluation metric to quantify language alignment, and (iii) demonstrate the impact of our approach across diverse multilingual NLP tasks.

To achieve these objectives, we propose a novel Cross-Lingual Mapping (CL) Task that explicitly models linguistic correspondences during the pre-training phase. Unlike existing approaches that rely on bilingual fine-tuning or contrastive learning, our method integrates bi-directional CL within the model’s learning process, facilitating improved alignment across languages. Additionally, we introduce the Language Alignment Coefficient (LAC), a new metric designed to evaluate the consistency of language representations in multilingual LLMs, particularly in low-resource scenarios.

\paragraph{Contributions}
This work makes the following key contributions: (i) We propose a CL Task that enhances language alignment during pre-training, allowing multilingual LLMs to learn direct cross-lingual correspondences. (ii) We introduce the LAC, a novel metric that quantifies cross-lingual consistency and provides a robust evaluation of multilingual representations. (iii) We construct an extensive evaluation framework encompassing MT, CLSum, CLQA and CLNLU, enabling a comprehensive assessment of multilingual performance. (iv) Our experiments demonstrate that the proposed approach significantly improves multilingual LLM performance, achieving up to 11.8 BLEU scores of improvement in MT, a 6.72 points increase in BERTScore-Precision in CLQA and more than a 5\% increase in CLNLU accuracy over strong baseline,, namely the Llama-3-8B model obtained by applying supervised fine-tuning (SFT) with the same instruction-tuning dataset.

These findings highlight the effectiveness of incorporating explicit cross-lingual objectives into pre-training, offering a promising direction for enhancing multilingual LLMs without sacrificing monolingual fluency.

\section{Related Work}
\label{sec:related work}
Improving language alignment in multilingual models is critical for enhancing cross-lingual performance, particularly for low-resource languages. Prior research has primarily focused on three strategies: parameter-sharing techniques, contrastive learning, and bilingual data integration during pre-training. While effective to some extent, these methods exhibit limitations that hinder robust cross-lingual generalization.

Parameter-sharing facilitates knowledge transfer by sharing model parameters and vocabularies across languages~\cite{muller2021alignpredictunderstandingcrosslingual, le2023bloom, conneau2019unsupervised}. This paradigm underlies many strong multilingual encoders such as InfoXLM~\cite{chi-etal-2021-infoxlm}, VECO~\cite{luo2021veco}, LaBSE~\cite{feng2022languageagnosticbertsentenceembedding}, and RemBERT~\cite{chuang2022diffcsedifferencebasedcontrastivelearning}, which jointly train a single model over shared subword vocabularies while injecting additional cross-lingual signals. However, it struggles with languages that have vastly different structures or limited training data. Another approach, masked token prediction on bilingual parallel sentences~\cite{NEURIPS2019_c04c19c2}, improves cross-lingual representations but is constrained by the availability of parallel corpora. 

Contrastive learning has also been widely explored for aligning multilingual representations. \citet{wang2021aligningcrosslingualsentencerepresentations} proposed aligning sentence embeddings via independent encoders, but these methods often accumulate errors, leading to instability. InfoXLM and LaBSE, for instance, apply contrastive or translation-ranking objectives on parallel sentences to obtain language-agnostic representations~\cite{chi-etal-2021-infoxlm,feng2022languageagnosticbertsentenceembedding}. Advances in unsupervised contrastive learning~\cite{gao2022simcsesimplecontrastivelearning, zhou2022debiasedcontrastivelearningunsupervised, chuang2022diffcsedifferencebasedcontrastivelearning, wu2020clearcontrastivelearningsentence} have improved alignment using data augmentation, yet they often require additional fine-tuning, which limits their generalization. Our approach circumvents these challenges by embedding cross-lingual alignment directly into pre-training, improving stability without post-training adjustments.

Bilingual data integration has shown promise in optimizing multilingual models. \citet{ham-kim-2021-semantic-alignment} employed a teacher-student framework for monolingual semantic alignment, while reinforcement learning methods~\cite{siddique2020unsupervised, gong2019reinforcement, chen2020text, yasui-etal-2019-using} used semantic similarity as a reward signal to improve language alignment. More recently, PreAlign~\cite{li-etal-2024-prealign} demonstrates that explicitly establishing multilingual alignment before large-scale language model pre-training can further boost cross-lingual transfer. However, these methods are computationally demanding and require fine-tuning stability. Similarly, bilingual pre-training techniques~\cite{kondo-etal-2024-enhancing, miao2024enhancing} and instruction fine-tuning~\cite{gao2024multilingualpretraininginstructiontuning} have demonstrated moderate success but remain insufficient for deep cross-lingual generalization.

Our work builds on these insights by introducing a novel CL task that enhances multilingual alignment at the pre-training stage. Unlike prior approaches that depend on post-hoc alignment techniques or large-scale bilingual corpora, our method embeds cross-lingual objectives directly into training, strengthening both multilingual generation and comprehension, providing a scalable and effective solution for improving multilingual LLMs.

\section{Language Alignment Coefficient (LAC)}
\label{sec:Language Alignment Evaluation Matrix}

Measuring language alignment in multilingual models is essential for evaluating their cross-lingual transferability. Existing methods primarily rely on the average cosine similarity of sentence embeddings across intermediate layers of LLMs~\cite{li2024quantifyingmultilingualperformancelarge}. Additionally, projection variance has been used to quantify alignment stability. However, these metrics often exhibit inconsistencies, as high cosine similarity does not necessarily imply robust alignment, particularly when test samples are small or susceptible to outliers. The variance in similarity scores across layers can lead to unstable assessments, limiting their reliability in evaluating multilingual representation alignment.

To overcome these limitations, we propose the \textit{Language Alignment Coefficient} (LAC), a novel metric that accounts for both alignment strength and stability. LAC is defined as the ratio of the average cosine similarity of sentence embeddings to the standard deviation of these similarity values across selected layers \( L_{\text{sub}} \). This formulation is statistically equivalent to the inverse of the \textit{Coefficient of Variation} (CV)~\cite{pearson1896vii}, ensuring a more stable and reliable measurement of alignment by incorporating both similarity magnitude and dispersion.

Let $X = \{x_k\}_{k=1}^n$ and $Y = \{y_k\}_{k=1}^m$ be sentence pairs in different languages, where $E_{i}(X)$ and $E_{i}(Y)$ denote their embeddings extracted from layer $i$ of the model by 
taking the hidden states of the last tokens $x_n$ and $y_m$, respectively. The LAC metric is defined as:

\begin{equation}
  \centering
  \label{eq:LAC}
  \text{LAC} = \frac{1}{|L_{\text{sub}}|} \sum_{i \in L_{\text{sub}}} \frac{\text{cos}(E_i(X), E_i(Y))}{\text{std}(\{\text{cos}(E_i(X), E_i(Y))\}_{i \in L_{\text{sub}}})},
\end{equation}

where $L_{\text{sub}} = \{5, 10, 15, 20, 25\}$ represents the subset of model layers used for evaluation, following prior work on multilingual alignment assessment~\cite{li2024quantifyingmultilingualperformancelarge}.

A higher LAC value indicates greater alignment stability, signifying that similarity scores remain consistent across different model layers. Unlike prior metrics that consider only mean similarity, LAC normalizes similarity scores by their dispersion, mitigating the effect of outliers and ensuring robust evaluation even in low-data multilingual scenarios. By integrating alignment strength and variability, LAC provides a more consistent metric for assessing multilingual LLMs, particularly in cases where test data is sparse or noisy.

\section{Language Alignment in Multilingual Large Language Models}
\label{methods}

\begin{figure}[t!]
    \centering
    \includegraphics[width=0.6\textwidth]{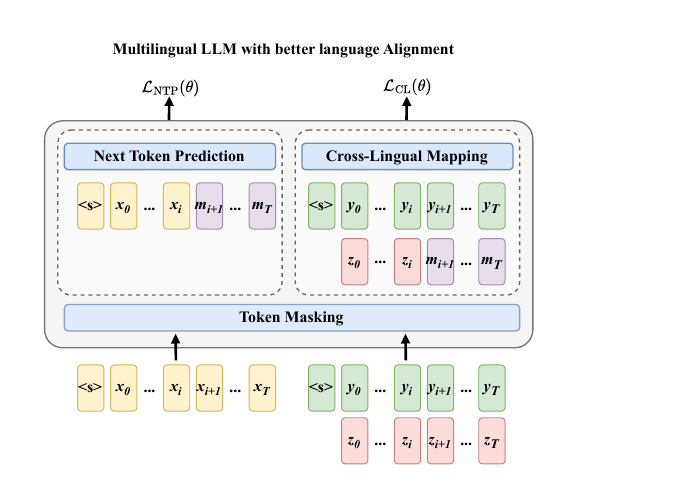}
    \caption{Continue pre-training including NTP and CL. "<s>" is "start token", "$m_i$" is the mask of the $i_{th}$ token. "x", "y", "z" are tokens from different languages.}
    \label{fig:flowchart}
\end{figure}

To address the challenges of language alignment in multilingual LLMs, we introduce a continued pre-training approach that simultaneously optimizes two objectives: the \textit{Next-Token Prediction} (NTP) task and the \textit{Cross-Lingual Mapping} (CL) task. These tasks are designed to enhance both the model’s monolingual generation capabilities and its ability to align representations across languages. The overall pre-training framework is illustrated in Figure~\ref{fig:flowchart}. In the NTP task, the model predicts the next token \( x_{i+1} \) based solely on the preceding tokens in the same language. In contrast,, in the CL task, the model generates a token \( z_{i+1} \) in a target language by using the complete sequence of the source language \( Y \) and previously generated target tokens. This dual-task approach ensures that multilingual alignment is achieved without sacrificing monolingual fluency.

\subsection{Next-Token Prediction Task}

A major challenge in training multilingual LLMs is the imbalance in pre-training data, which results in performance disparities between high-resource and low-resource languages. Prior studies have shown that improving low-resource language modeling enhances overall multilingual alignment~\cite{muller2021alignpredictunderstandingcrosslingual, xu2024paradigmshiftmachinetranslation}. However, continued pre-training on low-resource languages alone often leads to \textit{catastrophic forgetting} of previously learned high-resource languages~\cite{liu2019roberta, ke2022continual, gururangan2020don, mccloskey1989catastrophic, ratcliff1990connectionist}. 

To mitigate this issue, we adopt a decoder-only transformer model, parameterized by \( \theta \), for the NTP task. Given a monolingual sentence \( X = (x_0, x_1, \dots, x_T) \), the objective is to predict the next token \( x_t \) in the sequence, conditioning on the preceding tokens. This is formulated as a negative log-likelihood loss:

\begin{equation}
  \mathcal{L}_{\text{NTP}}(\theta) = - \sum_{t=0}^{T} \log P(x_t | x_0, \dots, x_{t-1}; \theta).
\end{equation}

This formulation ensures that the model maintains high-resource language proficiency while progressively improving its low-resource language capabilities. By training on a mixture of high- and low-resource languages, the model achieves better overall multilingual robustness without overfitting to dominant languages.

\subsection{Cross-Lingual Mapping Task}
\label{sec:Cross-Lingual Mapping Task}

Monolingual next-token prediction lacks explicit cross-lingual alignment signals. Without bilingual contexts, models trained on monolingual corpora reinforce same-language generation, limiting cross-lingual transfer in translation and multilingual understanding tasks.

To address this limitation, we introduce the  CL task, which explicitly trains the model to generate sequences in a target language while conditioning on a source language sequence. Given a bilingual parallel sentence pair \( (X, Y) \), where \( X \) is the source sentence and \( Y \) is the target sentence, the model learns to generate \( Y \) based on the information in \( X \). 

Prior work explored translation during pretraining under Masked Language Modeling (MLM)\cite{NEURIPS2019_c04c19c2}, and extensively in encoder-decoder architectures\cite{liu2019roberta}. However, MLM's bidirectional context differs fundamentally from Causal Language Modeling's left-to-right constraint, making cross-lingual alignment more challenging in decoder-only models. Research on effective cross-lingual alignment within CLM-based decoder-only architectures remains limited.

Additionally, LLMs differ from traditional models in their instruction-driven paradigm. Rather than fixed input formats, LLMs rely on prompts to determine task behavior.
Our approach differs from traditional translation-based pretraining in two ways:
(i) We enhance cross-lingual embedding similarity directly within decoder-only models, without explicit encoding-decoding separation;
(ii) We use no explicit task instructions, enabling generalizable, instruction-agnostic cross-lingual representations.
Unlike contrastive methods requiring explicit alignment, our CL task enables implicit transfer through a sequence-to-sequence objective:
\begin{equation}
\mathcal{L}{\text{CL}}(\theta) = -\sum{t=0}^{T} \log P(y_t | y_0, \dots, y_{t-1}; X; \theta).
\end{equation}
This ensures joint learning of semantic correspondences and syntactic transformations without predefined alignment constraints.

\subsection{Joint Optimization of Multilingual Alignment}

We jointly optimize the NTP and CL objectives during pre-training. The final loss function for continued pre-training is given by:

\begin{equation}
  \mathcal{L}_{\text{PT}}(\theta) = \mathcal{L}_{\text{NTP}}(\theta) + \mathcal{L}_{\text{CL}}(\theta).
\end{equation}

This dual-task training strategy enables the model to retain strong monolingual generation capabilities while simultaneously improving cross-lingual mapping. By jointly optimizing both objectives, the model learns to maintain high-resource language fluency while developing robust alignment mechanisms for low-resource languages. This approach results in a more effective multilingual LLM capable of both generating text fluently in a given language and transferring knowledge efficiently across linguistic boundaries.

\section{Dataset Creation}
\label{sec:Dataset Creation}
To ensure a comprehensive evaluation, we select the following language pairs: English-Chinese (EN-ZH), English-Czech (EN-CS), Chinese-Japanese (ZH-JP), and Czech-Ukrainian (CS-UK). These pairs represent both high- and low-resource scenarios, with EN-ZH (high-resource) and EN-CS (mid-resource) evaluating English-centric performance, while ZH-JP (high-resource) and CS-UK (low-resource) assess non-English alignment.

Our fine-tuning dataset is derived from \textit{Cleaned Alpaca}~\cite{alpaca}, augmented with machine translation (MT), cross-lingual question answering (CLQA), and cross-lingual summarization (CLSum) tasks to enhance the model’s multilingual capabilities. For CLSum, we utilize \textit{CrossSum}~\cite{bhattacharjee2021crosssum} for EN-ZH and ZH-JP, while CLQA data is sourced from \textit{OpenHermes-2.5}~\cite{OpenHermes-2.5}, supplemented with 2000 manually verified QA pairs (1000 EN-CS, 1000 CS-UK) generated using Claude 3.5. Evaluation is conducted using the XQuAD dataset~\cite{artetxe-etal-2020-cross} for CLQA and \textit{Belebele}~\cite{bandarkar-etal-2024-belebele} for CLNLU, focusing on our selected language pairs.

\begin{figure}[t!]
  \centering
  \includegraphics[width=00.9\textwidth]{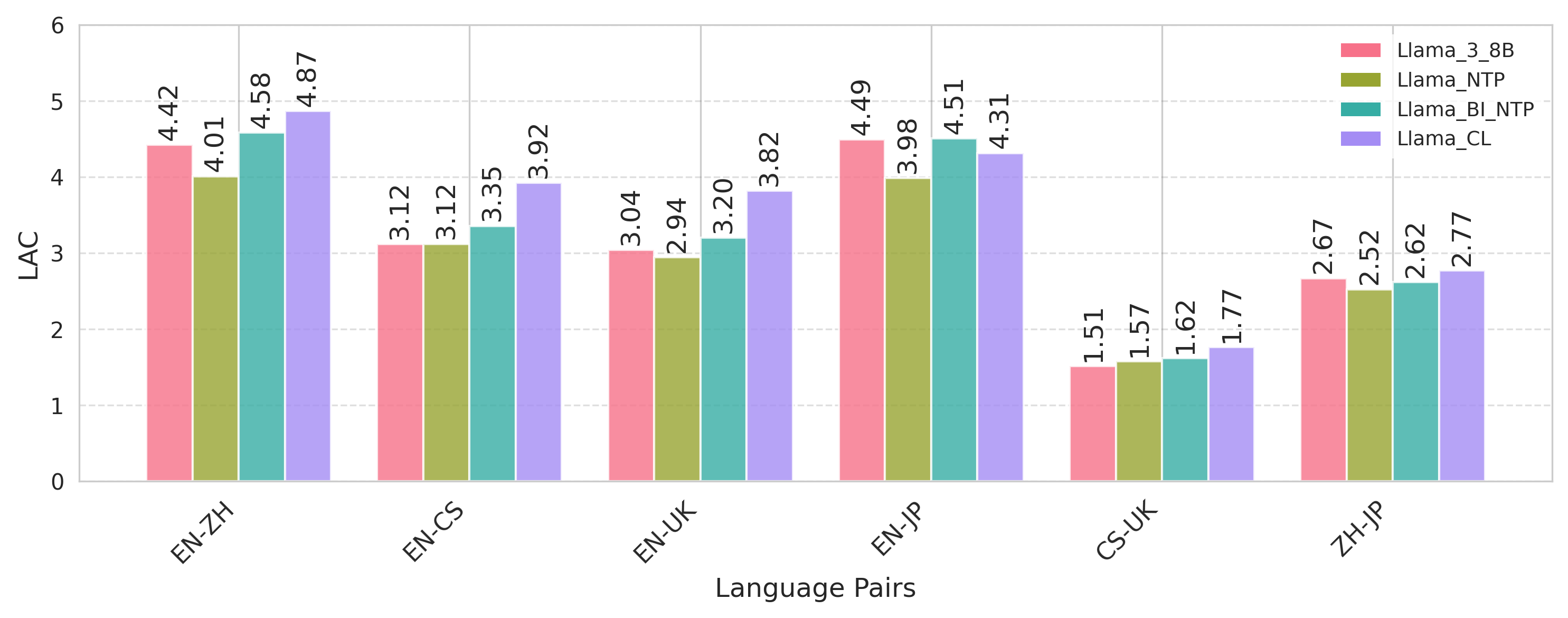}
  \caption{Language alignment of different pre-trained models. LAC is Language Alignment Coefficient.}
  \label{fig:similarity}
\end{figure}

Monolingual data for continued pre-training is sourced from \textit{CulturaX}~\cite{nguyen2023culturax}, covering English, Chinese, Czech, Ukrainian, and Japanese. English data is included to prevent catastrophic forgetting. Parallel data for MT and cross-lingual mapping tasks come from \textit{WMT2024}, with 50,000 sentence pairs per language pair for MT and 3.7 million pairs for cross-lingual mapping. During continued pre-training, we optimize two separate objectives: next-token prediction (NTP) and cross-lingual mapping. The NTP objective is trained on 4{,}128{,}308 instances ($\approx$4.1M), and the per-batch mixture allocates 52.7\% and 47.3\% of training examples to NTP and cross-lingual mapping, respectively. Since the two losses are of comparable magnitude, we do not apply additional task-specific weighting.
No dataset overlap occurs between training and evaluation phases.

Further details on dataset sources, language pairs, dataset sizes, and task-specific prompts are provided in Tables~\ref{tab:task_details} and~\ref{table:Prompts}.

\begin{table}[t!]
\centering
\small
\renewcommand{\arraystretch}{1.2}
\begin{tabular}{p{2.8cm}p{1.8cm}p{3cm}p{3.5cm}p{2.5cm}}
\hline
\textbf{Task Name} & \textbf{Sub-Task} & \textbf{\makecell[l]{Language /\\Language Pairs}} & \textbf{Data Volume} & \textbf{Data Source} \\
\hline
\multirow{5}{2.8cm}{Monolingual Continue Pretraining} & \multirow{5}{*}{NA} & English & 37,182,019 tokens & \multirow{5}{*}{ClutureX} \\
 & & Chinese & 93,124,828 tokens & \\
 & & Japanese & 85,415,308 tokens & \\
 & & Ukrainian & 187,336,165 tokens & \\
 & & Czech & 185,235,692 tokens & \\
\hline
\makecell[l]{Cross-Lingual \\Mapping} & NA & \makecell[l]{EN-ZH, EN-CS,\\CS-UK, ZH-JP} & \makecell[l]{3.7M sentence\\pairs per language pair} & WMT2024 \\
\hline
\multirow{4}{2.8cm}{Instruction\\Finetuning for\\general task} & MT & \makecell[l]{EN-ZH, EN-CS,\\CS-UK, ZH-JP} & \makecell[l]{50K per language pair} & WMT2024 \\
 & CLQA & \makecell[l]{EN-CS \& CS-UK} & \makecell[l]{1000 per language pair} & \makecell[l]{Tranlate from\\OpenHermes-2.5} \\
 & \multirow{2}{*}{\makecell[l]{CL-Sum}} & EN-ZH & 3,893 & \multirow{2}{*}{\makecell[l]{CrossSum}} \\
 & & ZH-JP & 809 & \\
\hline
\multirow{9}{2.8cm}{Evaluation} & \multirow{4}{1.8cm}{\makecell[l]{MT}} & \multirow{4}{2.5cm}{\makecell[l]{EN-ZH, EN-CS,\\CS-UK, ZH-JP}} & \makecell[l]{FLORES-200:\\1012 per language pair} & \multirow{4}{2cm}{\makecell[l]{FLORES-200,\\WMT 2022}} \\
 & & & \makecell[l]{WMT cs-uk: 1930} & \\
 & & & \makecell[l]{WMT en-cs: 2037} & \\
 & & & \makecell[l]{WMT en-zh: 2037} & \\[1ex]
 & Xquad & EN-ZH & 1190 & Xquad \\[1ex]
 & \makecell[l]{Open-End\\CLQA} & \makecell[l]{EN-CS \& CS-UK} & \makecell[l]{500 per language pair} & \makecell[l]{Tranlate from\\OpenHermes-2.5} \\
 & \multirow{2}{*}{\makecell[l]{CL-Sum}} & EN-ZH & 101 & \multirow{2}{*}{\makecell[l]{CrossSum}}\\
 & & ZH-JP & 486 & \\
 & \makecell[l]{CLNLU} & \makecell[l]{EN-ZH, EN-CS,\\CS-UK, ZH-JP} & \makecell[l]{900 per language pair} & \makecell[l]{Belebele} \\
\hline
\end{tabular}
\caption{Task Details and Data Statistics}
\label{tab:task_details}
\end{table}

\begin{table}[t!]
\small
\centering
\begin{tabular}{@{}p{6cm} p{8cm}@{}}
\toprule
\textbf{Task} & \textbf{Prompts} \\ 
\midrule
MT & Translate the following <Source Language> to <Target Language>: <Source Sentence> \\ \cline{2-2}
CLSum & Please read the following <Source Language> text and generate a short and precise <Target Language> summary: <Source Language Paragraph> \\ \cline{2-2}
Xquad & <Source Language Paragraph> Based on the above paragraph, answer the following questions in <Target Language> or numbers. \\ \cline{2-2}
Cross-Lingual Open-ended Question and Answering & Answer the following questions in <Target Language>. <Question in Source Language> \\  \cline{2-2}
CLNLU & Answer the following single-choice questions and output only the option. <Source Question> <Target Options>. \\
\bottomrule
\end{tabular}
\caption{Prompts for Different Tasks}
\label{table:Prompts}
\end{table}

\section{Experiments}

\subsection{Experiment Settings}
We use Llama-3-8B~\cite{dubey2024llama3herdmodels} as the base model, given its extensive multilingual capabilities across 31 languages. Continued pre-training is performed on Llama-3-8B, followed by fine-tuning on both cross-lingual and monolingual tasks. To evaluate language alignment and generation quality, we compute sentence embedding similarities and perplexity scores. For instruction fine-tuning, we train models on general and cross-lingual tasks to assess downstream performance. Given that MT is the most representative cross-lingual task, ablation studies focus on it to isolate key factors influencing cross-lingual effectiveness. 

\subsection{Training details}
\label{training details}

We conducted training with a warm-up ratio of 0.01 and a sequence length of 2048 tokens, limiting the process to 1 epoch for pre-training and 3 epochs for general fine-tuning. The training employed bf16 precision with Low-Rank Adaptation (LoRA)~\cite{hu2021loralowrankadaptationlarge}, configured with a LoRA rank of 16 and target modules applied to all layers. We utilized 8 H100 GPUs, with each GPU processing 16 batches and an 8-step gradient accumulation, yielding an effective batch size of 1024. The initial learning rate was set to 1e-4 with the AdamW optimizer. Under this configuration, continued pre-training required approximately 70 hours, while fine-tuning was completed in 53 minutes.

\subsection{Language Alignment and Perplexity Evaluation of Base Models}
\label{sec:base_model_eval}

We assess language alignment and monolingual perplexity on Llama-3-8B and its variants following different pre-training strategies:
\begin{CJK*}{UTF8}{gbsn}
\begin{enumerate}
    \setlength{\itemsep}{0pt}
    \setlength{\parskip}{0pt}
    \item \textbf{Llama\_NTP:} Trained with monolingual next-token prediction.
    \item \textbf{Llama\_Bi\_NTP:} The model is trained on a hybrid corpus comprising both monolingual data and bilingual sentence pairs~\cite{kondo-etal-2024-enhancing}. While maintaining the standard next-token prediction objective, the training data encompasses both monolingual sequences and concatenated bilingual sentence pairs (e.g., "How are you? 你好吗？"). This training configuration serves to augment the model's cross-lingual transition probability.
    \item \textbf{Llama\_CLI:} The model is jointly trained on monolingual next-token prediction and translation tasks during the pre-training stage, where translation instances are formulated with an explicit instruction prompt, e.g., ``Translate the following {language 1} to {language 2}.''
    \item \textbf{Llama\_CL:} Our proposed model, incorporating monolingual next-token prediction and cross-lingual mapping.
\end{enumerate}
\end{CJK*}

The CL objective is designed to align cross-lingual semantic spaces by predicting target-language tokens from source-language sentence embeddings. While Bi\_NTP remains fundamentally a Next-Token Prediction (NTP) task, it extends standard monolingual NTP by incorporating code-switched sequences constructed from concatenated bilingual parallel sentences. Under this configuration, the model primarily learns the mechanisms of linguistic transition. Although CL conceptually overlaps with machine translation, its scope is broader; by aligning latent semantic spaces, it provides a robust foundation for a wide range of cross-lingual downstream applications beyond simple translation. We evaluate bilingual sentence alignment using Flores-200 and WMT2022, while monolingual perplexity is tested on 1,000 samples per language from CulturaX~\cite{nguyen2023culturax}.

Results in Table~\ref{tab:perplexity-scores} and Figure~\ref{fig:similarity} show that while Llama\_NTP and Llama\_Bi\_NTP significantly reduce monolingual perplexity, particularly in low-resource languages like CS (-4.33 points), their improvements in cross-lingual alignment remain inconsistent. Llama\_Bi\_NTP achieves gains primarily for CS-UK but shows limited impact on EN-CS due to the lack of an explicit CL objective.

In contrast, Llama\_CL achieves a more balanced improvement across both monolingual perplexity and cross-lingual alignment. The integration of CL objective explicitly reinforces bilingual relationships, leading to notable performance advantages: a 0.57-point improvement over Llama\_Bi\_NTP for EN-CS and consistent gains for EN-ZH (0.2), CS-UK (0.15), and ZH-JP (0.15). 

Further, Llama\_CL demonstrates strong generalization to unseen language pairs, outperforming other models on EN-UK alignment by 0.77, 0.87, and 0.62 points against Llama-3-8B, Llama\_NTP, and Llama\_Bi\_NTP, respectively, while also achieving stable alignment for EN-JP. This suggests that our method is able to generalized beyond trained bilingual pairs, reinforcing multilingual robustness.

Overall, these findings highlight the necessity of explicit cross-lingual mapping in pre-training. While NTP enhances monolingual fluency, our approach successfully bridges linguistic gaps, making it particularly effective for low-resource language processing and multilingual applications.

\begin{table}[t!]
\small
  \centering
  \begin{tabular}{lBBBBB}
    \toprule
    \textbf{Lang} & \textbf{L-3-8B} & \textbf{L\_NTP} & \textbf{L\_Bi\_NTP} & \textbf{L\_CL-only} & \textbf{L\_CL} \\
    \midrule
    EN & 30.1 & 27.4 & 26.9 & 30.7 & \textbf{25.5} \\
    ZH & 22.4 & 22.1 & 21.3 & 23.3 & \textbf{20.6} \\
    CS & 21.1 & 16.8 & 16.1 & 22.1 & \textbf{15.2} \\
    UK & 19.1 & 17.2 & 16.9 & 19.5 & \textbf{16.0} \\
    JP & 25.1 & 22.9 & 22.2 & 25.5 & \textbf{20.8} \\
    \bottomrule
  \end{tabular}
\caption{Perplexity Scores of Languages for Different Models. Best scores are in \textbf{bold}.}
\label{tab:perplexity-scores}
\end{table}

\begin{table}[t!]
\small
\centering
\setlength{\tabcolsep}{2pt}
\begin{tabular}{llclcccc}
\toprule
\textbf{Tasks} & \textbf{Data Sets} & \textbf{Metric} & \textbf{Llama-3-8B} & \textbf{Llama\_NTP} & \textbf{Llama\_Bi\_NTP} & \textbf{Llama\_CLI} & \textbf{Llama\_CL} \\
\midrule
\multirow{14}{*}{MT} 
  & WMT-EN-ZH     & \multirow{7}{*}{BLEU} 
    & \varcell{36.2$\pm$0.5}
    & \varcell{37.8$\pm$0.5 (+1.6)}
    & \varcell{44.0$\pm$0.5 (+7.8)}
    & \varbest{49.4$\pm$0.5 (+13.1)}
    & \varcell{48.2$\pm$0.6 (+12.0)} \\
  & WMT-EN-CS     
    & 
    & \varcell{19.4$\pm$0.3}
    & \varcell{25.1$\pm$0.4 (+5.7)}
    & \varcell{26.3$\pm$0.4 (+6.9)}
    & \varbest{31.4$\pm$0.6 (+12.0)}
    & \varcell{31.2$\pm$0.5 (+11.8)} \\
  & WMT-CS-UK     
    & 
    & \varcell{18.2$\pm$0.3}
    & \varcell{21.0$\pm$0.4 (+2.8)}
    & \varcell{23.4$\pm$0.4 (+5.2)}
    & \varcell{25.1$\pm$0.4 (+6.9)}
    & \varbest{25.8$\pm$0.5 (+7.6)} \\
  & Flores-EN-ZH 
    & 
    & \varcell{28.6$\pm$0.4}
    & \varcell{29.9$\pm$0.4 (+1.3)}
    & \varcell{33.6$\pm$0.5 (+5.0)}
    & \varbest{37.6$\pm$0.4 (+9.0)}
    & \varcell{36.2$\pm$0.5 (+7.6)} \\
  & Flores-EN-CS 
    & 
    & \varcell{17.2$\pm$0.3}
    & \varcell{21.0$\pm$0.4 (+3.8)}
    & \varcell{21.6$\pm$0.4 (+4.4)}
    & \varbest{25.7$\pm$0.5 (+8.5)}
    & \varcell{23.2$\pm$0.4 (+6.0)} \\
  & Flores-CS-UK 
    & 
    & \varcell{14.3$\pm$0.3}
    & \varcell{18.0$\pm$0.3 (+3.7)}
    & \varcell{18.5$\pm$0.4 (+4.2)}
    & \varbest{20.0$\pm$0.3 (+5.7)}
    & \varcell{19.4$\pm$0.4 (+5.1)} \\
  & Flores-ZH-JP 
    & 
    & \varcell{15.7$\pm$0.3}
    & \varcell{17.4$\pm$0.3 (+1.7)}
    & \varcell{18.5$\pm$0.4 (+2.8)}
    & \varbest{21.6$\pm$0.4 (+5.9)}
    & \varcell{21.0$\pm$0.4 (+5.3)} \\
\cmidrule(lr){2-8}
  & WMT-EN-ZH     & \multirow{7}{*}{COMET} 
    & \varcell{67.31$\pm$0.29}
    & \varcell{70.30$\pm$0.33 (+2.99)}
    & \varcell{70.90$\pm$0.34 (+3.59)}
    & \varbest{72.34$\pm$0.21 (+5.03)}
    & \varcell{71.10$\pm$0.35 (+3.79)} \\
  & WMT-EN-CS     
    & 
    & \varcell{64.39$\pm$0.28}
    & \varcell{67.40$\pm$0.32 (+3.01)}
    & \varcell{70.10$\pm$0.36 (+5.71)}
    & \varbest{73.56$\pm$0.39 (+9.17)}
    & \varcell{72.60$\pm$0.40 (+8.21)} \\
  & WMT-CS-UK     
    & 
    & \varcell{75.33$\pm$0.31}
    & \varcell{82.60$\pm$0.38 (+7.27)}
    & \varcell{84.00$\pm$0.39 (+8.67)}
    & \varbest{86.28$\pm$0.42 (+10.95)}
    & \varcell{85.10$\pm$0.41 (+9.77)} \\
  & Flores-EN-ZH 
    & 
    & \varcell{75.74$\pm$0.31}
    & \varcell{82.90$\pm$0.38 (+7.16)}
    & \varcell{82.80$\pm$0.38 (+7.06)}
    & \varbest{83.90$\pm$0.11 (+8.16)}
    & \varcell{83.20$\pm$0.39 (+7.46)} \\
  & Flores-EN-CS 
    & 
    & \varcell{75.18$\pm$0.30}
    & \varcell{81.20$\pm$0.36 (+6.02)}
    & \varcell{83.30$\pm$0.39 (+8.12)}
    & \varbest{88.03$\pm$0.23 (+12.85)}
    & \varcell{85.00$\pm$0.42 (+9.82)} \\
  & Flores-CS-UK 
    & 
    & \varcell{76.47$\pm$0.30}
    & \varcell{84.30$\pm$0.39 (+7.83)}
    & \varcell{84.90$\pm$0.40 (+8.43)}
    & \varbest{87.46$\pm$0.38 (+10.99)}
    & \varcell{86.30$\pm$0.43 (+9.83)} \\
  & Flores-ZH-JP 
    & 
    & \varcell{81.63$\pm$0.32}
    & \varcell{84.60$\pm$0.35 (+2.97)}
    & \varcell{85.90$\pm$0.37 (+4.27)}
    & \varbest{88.99$\pm$0.30 (+7.36)}
    & \varcell{87.60$\pm$0.40 (+5.97)} \\
\midrule
\multirow{4}{*}{CLSum} 
  & \multirow{3}{*}{CrossSum} & R-1  
    & 15.95 & 14.03(-1.92) & 15.52(-0.43) & 13.27(-2.68) & \textbf{16.80(+0.85)} \\
  &                            & R-2  
    & \textbf{5.964} & 4.992(-0.972) & 4.824(-1.140) & 4.329(-1.635) & 5.937(-0.027) \\
  &                            & R-L  
    & 15.77 & 13.91(-1.86) & 15.34(-0.43) & 12.78(-2.99) & \textbf{16.39(+0.62)} \\
  &                            & Rec 
    & 73.4 & 73.24(-0.16) & 73.09(-0.31) & 71.52(-1.88) & \textbf{74.12(+0.72)} \\
\midrule
\multirow{3}{*}{CLQA}  
  & \multirow{2}{*}{Xquad}   & F1              
    & \varcell{86.80$\pm$0.27}
    & \varcell{87.50$\pm$0.28 (+0.70)}
    & \varbest{88.70$\pm$0.30 (+1.90)}
    & \varcell{85.71$\pm$0.38 (-1.09)}
    & \varcell{88.50$\pm$0.30 (+1.70)} \\
  &                            & EM 
    & 17.48 & 18.82(+1.34) & 19.16(+1.68) & 16.35(-1.13) & \textbf{19.24(+1.76)} \\
  &                            & Prec 
    & 80.54 & 86.28(+5.74)  & 86.78(+6.24) & 79.33(-1.21) & \textbf{87.26(+6.72)} \\
\midrule
\multirow{4}{*}{CLNLU} 
  & Belebele-EN-ZH & \multirow{4}{*}{Acc} 
    & 58.78\% & 61.67\%(+2.89\%) & \textbf{62.89\%(+4.11\%)} & 61.33\%(+2.55\%) & 61.78\%(+3.00\%) \\ 
  & Belebele-EN-CS & 
    & 55.67\% & 59.78\%(+4.11\%) & 60.78\%(+5.11\%) & 58.56\%(+2.89\%) & \textbf{61.33\%(+5.66\%)} \\ 
  & Belebele-CS-UK & 
    & 47.56\% & 49.78\%(+2.22\%) & 51.33\%(+3.77\%) & 47.02\%(-0.54\%) & \textbf{54.22\%(+6.66\%)} \\ 
  & Belebele-ZH-JP & 
    & 40.22\% & 42.44\%(+2.22\%) & 45.67\%(+5.45\%) & 43.85\%(+3.63\%) & \textbf{46.56\%(+6.34\%)} \\
\bottomrule
\end{tabular}
\caption{Evaluation Results on Cross-Lingual Tasks for Llama-3. All results in this table are obtained by evaluating the instruction-fine-tuned models.} R-1, R-2, R-L, Rec, EM, Prec, and Acc respectively represent ROUGE-1, ROUGE-2, ROUGE-L, BERTScore-Recall, Exact Match, BERTScore-Precision, and Accuracy. Best scores are in \textbf{bold}.
\label{table:cross_lingual_result}
\end{table}

\subsection{Cross-Lingual Task Evaluation of Instruction-Tuned Models}
\label{sec:evaluation on tuned model}

\begin{table}[t!]
\small
\centering
\begin{tabular}{llclccc}
\toprule
\textbf{Tasks} & \textbf{Data Sets} & \textbf{Metric} & \textbf{BLOOM-3B} & \textbf{BLOOM\_NTP} & \textbf{BLOOM\_Bi\_NTP} & \textbf{BLOOM\_CL} \\
\midrule
\multirow{4}{*}{MT} 
  & WMT-EN-ZH    & \multirow{2}{*}{BLEU} 
    & \varcell{11.3$\pm$0.4}
    & \varcell{13.6$\pm$0.4 (+2.3)}
    & \varcell{16.6$\pm$0.4 (+5.3)}
    & \varbest{17.5$\pm$0.5 (+6.2)} \\
  & Flores-EN-ZH &                         
    & \varcell{13.0$\pm$0.4}
    & \varcell{16.7$\pm$0.4 (+3.7)}
    & \varcell{18.9$\pm$0.5 (+5.9)}
    & \varbest{22.6$\pm$0.5 (+9.6)} \\
\cmidrule(lr){2-7}
  & WMT-EN-ZH    & \multirow{2}{*}{COMET} 
    & \varcell{57.93$\pm$0.28}
    & \varcell{58.49$\pm$0.30 (+0.56)}
    & \varbest{59.82$\pm$0.34 (+1.89)}
    & \varcell{59.70$\pm$0.34 (+1.77)} \\
  & Flores-EN-ZH &                         
    & \varcell{60.25$\pm$0.30}
    & \varcell{62.63$\pm$0.33 (+2.38)}
    & \varcell{70.41$\pm$0.40 (+10.16)}
    & \varbest{74.68$\pm$0.43 (+14.43)} \\
\midrule
\multirow{4}{*}{CLSum} 
  & \multirow{3}{*}{CrossSum-EN-ZH} & R-1  
    & 6.20 & 6.70(+0.5) & 7.10(+0.9) & \textbf{7.30(+1.1)} \\
  &                                   & R-2  
    & 1.90 & 2.20(+0.3) & 2.90(+1.0) & \textbf{3.40(+1.5)} \\
  &                                   & R-L  
    & 5.12 & 5.43(+0.31) & 5.71(+0.59) & \textbf{6.88(+1.76)} \\
  &                                   & Rec 
    & 62.70 & 63.12(+0.42) & 63.89(+1.19) & \textbf{64.13(+1.43)} \\
\midrule
\multirow{3}{*}{CLQA}  
  & \multirow{3}{*}{Xquad-EN-ZH}    & F1 
    & \varcell{24.42$\pm$0.25}
    & \varcell{25.73$\pm$0.27 (+1.31)}
    & \varcell{27.08$\pm$0.28 (+2.66)}
    & \varbest{27.31$\pm$0.29 (+2.89)} \\
  &                                   & EM 
    & 8.24 & 8.82(+0.58) & 9.24(+1.00) & \textbf{9.41(+1.17)} \\
  &                                   & Prec 
    & 37.35 & 39.97(+2.62) & 42.18(+4.83) & \textbf{43.04(+5.69)} \\
\midrule
\multirow{1}{*}{CLNLU} 
  & Belebele-EN-ZH                  & Acc 
    & 41.11\% & 43.11\%(+2.00\%) & 44.67\%(+3.56\%) & \textbf{46.33\%(+5.22\%)} \\
\bottomrule
\end{tabular}
\caption{Evaluation Results on Cross-Lingual Tasks for BLOOM. All results in this table are obtained by evaluating the instruction-fine-tuned models.} R-1, R-2, R-L, Rec, EM, Prec, and Acc respectively represent ROUGE-1, ROUGE-2, ROUGE-L, BERTScore-Recall, Exact Match, BERTScore-Precision, and Accuracy. Best scores are in \textbf{bold}.
\label{table:cross_lingual_result_for_BLOOM}
\end{table}

To ensure a fair comparison and isolate the effectiveness of our pre-training strategy, we applied an identical instruction fine-tuning (IFT) process to all model variants described in Section~\ref{sec:base_model_eval}. The set of models includes the Llama-3-8B base model, baseline variants (Llama\_NTP and Llama\_Bi\_NTP), a joint-objective baseline (Llama\_CLI, trained with next-token prediction and translation objectives), and our proposed model, Llama\_CL. Consequently, all downstream performance gains reported in this section represent comparisons among fine-tuned checkpoints. We evaluated their performance on MT, CLSum, CLQA and CLNLU. For CLSum, we used the \textit{CrossSum} dataset~\cite{bhattacharjee2021crosssum} for EN-ZH and ZH-JP summarization. CLQA evaluation consisted of two components: reference-based QA using XQuAD~\cite{artetxe-etal-2020-cross} (EN-ZH) and open QA, constructed from \textit{OpenHermes-2.5}~\cite{OpenHermes-2.5}. And Belebele~\cite{bandarkar-etal-2024-belebele} is used as the CLNLU evaluation dataset. Reference evaluation results are presented in Table~\ref{table:cross_lingual_result}. 

To assess generalizability, we replicated these experiments on BLOOM-3B~\cite{le2023bloom}, observing similar trends to Llama-3-8B. Since many of the languages in our test set are not supported by BLOOM, we have chosen to report scores only for the EN-ZH language pair on the BLOOM-3B models. This ensures the validity and fairness of our evaluation.

Similar to the results obtained on the Llama-3-8B model, as shown in Table~\ref{table:cross_lingual_result_for_BLOOM}, our approach also demonstrates significant improvements in translation tasks on the BLOOM-3B model. In particular, on the Flores-EN-ZH dataset, our method achieves an impressive gain of 14.43, highlighting its effectiveness.

For the CLQA task, while the Llama-3-8B model slightly lags behind the Llama\_Bi\_NTP model in F1 score, our method consistently outperforms other approaches on the BLOOM-3B model, showcasing significant improvements over the baseline. Notably, it achieves a 2.89\% increase in F1, a 1.17\% gain in Exact Match Rate, and a 5.69\% boost in BERTScore-Precision.

Surprisingly, our method demonstrated greater improvements on the CLSum task with the Bloom-3B model compared to Llama-3-8B, achieving over a 1-point increase in ROUGE-1, ROUGE-2, ROUGE-L, and BERTScore-Precision. While there is still room for further enhancement, these results are promising. Additionally, our approach also delivered significant improvements on the CLNLU task with the BLOOM model.

By comparing Table~\ref{table:cross_lingual_result} and Table~\ref{table:cross_lingual_result_for_BLOOM}, it becomes clear that our method yields even greater improvements when applied to a base model with weaker foundational capabilities.

\subsubsection{Reference-based Evaluation}
\label{sec:Reference-based Evaluation}

Table~\ref{table:cross_lingual_result} presents a comparative analysis of our proposed model (Llama\_CL) against the Llama-3-8B baseline and several cross-lingual variants. The results indicate that while all adaptation strategies improve upon the vanilla baseline in Machine Translation (MT) and CLNLU, they exhibit distinct trade-offs in terms of task-specific performance versus general-purpose cross-lingual capability.

Specifically, the basic variants Llama\_NTP and Llama\_Bi\_NTP demonstrate steady improvements in translation and understanding. The competitive performance of Llama\_Bi\_NTP in CLQA (e.g., 88.70 F1 on Xquad) suggests that bilingual Next Token Prediction (NTP) enhances the model's capacity to perform language transitions at contextually appropriate positions, thereby providing a solid foundation for cross-lingual alignment. However, a more complex pattern emerges when considering Llama\_CLI, which incorporates explicit translation instructions during the pre-training phase. While Llama\_CLI achieves the highest performance in most MT benchmarks—surpassing the baseline by up to 13.1 BLEU points, it suffers from significant performance degradation in CLSum and CLQA. For instance, in CLSum, Llama\_CLI experiences a 2.68-point drop in ROUGE-1 compared to the baseline.

These results suggest that retaining task-specific instructions during the pre-training stage can easily trap the model in a task-specific local optimum. Although such a strategy may yield marginal gains in certain downstream tasks like CLNLU, it appears to induce a degree of overfitting to the translation objective that even subsequent fine-tuning cannot fully rectify. This rigid optimization limits the model’s flexibility in handling complex, open-ended cross-lingual reasoning tasks like summarization.

In contrast, our approach (Llama\_CL) effectively mitigates this risk. By balancing instruction-tuning with robust cross-lingual mapping, Llama\_CL achieves MT performance comparable to the specialized Llama\_CLI while remaining the only variant to consistently outperform the baseline in CLSum. Furthermore, in CLQA, Llama\_CL attains the highest Exact Match (19.24\%) and BERTScore-Precision (87.26) scores. These findings underscore that our method fosters a more versatile cross-lingual representation space, ensuring high-fidelity generation and comprehension without sacrificing the model's generalizability across diverse task formats.

\subsubsection{LLM-Assisted Reference-Free Evaluation}
\label{sec:open QA analysis}

To further assess cross-lingual generalization, we evaluated models on open-domain question answering, where responses must be coherent and contextually relevant without explicit references. We constructed CLQA datasets for EN-CS and CS-UK using a subset of open-ended questions from \textit{OpenHermes-2.5}~\cite{OpenHermes-2.5}. Common question types and examples are provided in Table~\ref{tab:open-ended-questions}. Given the subjective nature of queries such as \textit{“Could you share a library-related joke?”}, responses were rated on a 1–4 scale using Claude 3.5, with a `\textit{Same}' rating assigned when models produced indistinguishable or uniformly incorrect answers.

\begin{table}[t!]
\centering
\small
\begin{tabular}{@{}p{6cm}p{8cm}@{}}
\toprule
\textbf{Types of open-ended questions} & \textbf{Question Example} \\ \midrule
Creative prompts & Compose a poem about childhood. \\
Commonsense reasoning & Why do people wear sunscreen? \\
Knowledge-based queries & What are the key features of quantum computing? \\
List-style questions & List five characteristics of a good employee. \\ \bottomrule
\end{tabular}
\caption{Examples of open-ended question types}
\label{tab:open-ended-questions}
\end{table}

\begin{table}[t!]
\small
\centering
\begin{tabular}{ccccc}
\hline
\textbf{Rank} & \textbf{L-3-8B} & \textbf{L\_NTP} & \textbf{L\_Bi\_NTP} & \textbf{L\_CL} \\ 
\hline
1   & 181   & 271   & 252   & \textbf{325}   \\ 
2   & 332   & 182   & 281   & 190   \\ 
3   & 261   & 213   & 212   & 202   \\ 
4   & 163   & 271   & 192   & 220   \\ 
\hline
\textbf{Same} & \multicolumn{4}{c}{63}      \\ 
\hline
\end{tabular}
\caption{Evaluation Results on Open-Ended CLQA Task. Best scores are in \textbf{bold}.}
\label{table:open-end-analysis}
\end{table}

As shown in Table~\ref{table:open-end-analysis}, Llama\_CL achieved the highest performance, receiving 325 top ratings out of 1,000 responses. It consistently outperformed other models, particularly in tasks requiring nuanced comprehension and complex text generation. Notably, when prompted to \textit{“Compose a poem about childhood”}, other models provided basic descriptions, whereas Llama\_CL employed advanced poetic techniques such as parallelism, demonstrating improved contextual understanding.

These results highlight the benefits of cross-lingual mapping in enhancing multilingual generative capabilities. However, all models exhibited weaknesses in complex reasoning tasks, underscoring the need for further improvements in logical inference and cross-lingual reasoning.

\section{Ablation Study}
\label{sec: ab study}

\subsection{Analysis of Cross-Lingual Training Objectives}
We conducted ablation experiments on MT to isolate key factors influencing model performance. The experiments were structured as follows:

\begin{enumerate}
    \setlength{\itemsep}{0pt}
    \setlength{\parskip}{0pt}
    \item \textbf{$E_{sep}$}: Bilingual data was split into monolingual sentences and used for next-token prediction.
    \item \textbf{$E_{post\_mt}$}: Bilingual data was converted into instruction-tuning format and used during fine-tuning with the prompt: ``\textit{Translate the following \textless Source Language\textgreater\ sentence into \textless Target Language\textgreater.}''
    \item \textbf{$E_{pre\_mt}$}: Bilingual data was incorporated as instruction-tuning data during continued pre-training using the same translation prompt as $E_{post\_mt}$.
    \item \textbf{$E_{cross}$}: Our proposed approach, integrating cross-lingual mapping into pre-training.
\end{enumerate}

As shown in Table~\ref{tab:comet_scores}, the comparison between $E_{sep}$ and $E_{cross}$ confirms that improved cross-lingual performance is not merely a result of increased data volume. The significant performance gap between $E_{post\_mt}$ and $E_{cross}$ underscores the importance of embedding cross-lingual tasks during pre-training, as instruction fine-tuning alone provides limited benefits.

While $E_{pre\_mt}$ and $E_{cross}$ achieve comparable results, on average, $E_{cross}$ outperforms $E_{pre_mt}$ by 2.37 COMET points across the seven language pairs, with the largest margin observed on the CS-UK pair (3.8 points), indicating stronger language alignment when cross-lingual tasks are explicitly modeled during pre-training. This suggests that premature instruction fine-tuning, as in $E_{pre\_mt}$, may lead to overfitting, whereas $E_{cross}$ facilitates better generalization across language pairs.

These findings highlight the necessity of cross-lingual mapping in pre-training for establishing a robust multilingual foundation. Explicitly integrating cross-lingual objectives during pre-training enables the model to develop deeper alignment, resulting in improved transferability across languages compared to post-hoc instruction fine-tuning.

\begin{table}[t!]
\small
  \centering
  \begin{tabular}{lcccc}
    \toprule
    \textbf{Test Datasets} & \textbf{$E\_{sep}$} & \textbf{$E\_{post\_mt}$} & \textbf{$E\_{pre\_mt}$} & \textbf{$E\_{cross}$} \\
    \midrule
    WMT-en-zh    & 62.04 & 67.26 & 68.87 & 71.03 \\
    WMT-en-cs    & 66.72 & 63.42 & 69.78 & 72.47 \\
    WMT-cs-uk    & 77.92 & 74.21 & 81.17 & 84.97 \\
    Flores-en-zh & 75.64 & 78.31 & 81.09 & 83.05 \\
    Flores-en-cs & 79.95 & 76.44 & 83.02 & 84.87 \\
    Flores-cs-uk & 79.96 & 60.11 & 83.45 & 86.04 \\
    Flores-zh-jp & 76.21 & 71.45 & 85.97 & 87.51 \\
    \bottomrule
  \end{tabular}
  \caption{COMET score for MT task of models based on different pre-training settings. E\_cross is our method.}
  \label{tab:comet_scores}
\end{table}

\subsection{Performance on English Monolingual Downstream Tasks}
\label{sec: English_Performance}

Furthermore, we compared the results of fine-tuning Llama-3-8B (L-3-8B-Ins) and Llama\_CL (L\_CL\_Ins) directly using the Alpaca dataset~\cite{alpaca} on English-only downstream tasks. For each base model, we conducted LoRA fine-tuning for two epochs with a rank of 16, an initial learning rate of $1e^{-4}$, a warm-up rate of 0.1, and a batch size of 256. We selected MMLU~\cite{hendrycks2021measuringmassivemultitasklanguage}, LogiQA~\cite{liu2020logiqachallengedatasetmachine}, and AlpacaEval~\cite{alpaca_eval} to evaluate the models' capabilities in language understanding, logical reasoning, and open-ended generation, respectively. 

As shown in Table~\ref{tab:english_performance}, the model trained with our method achieves performance on downstream English tasks that is comparable to the original base model, and even slightly surpasses it on the MMLU benchmark. These results suggest that our pre-training strategy enhances cross-lingual transfer while maintaining English monolingual proficiency with negligible degradation. We attribute this to retaining the Next-Token Prediction (NTP) objective and preserving a portion of English monolingual data during continued pre-training, which together help mitigate catastrophic forgetting.

\begin{table}[t!]
\small
    \centering
    \begin{tabular}{lccc} 
        \toprule
        \textbf{Model} & \textbf{MMLU} & \textbf{LogiQA} & \textbf{Alpaca\_Eval} \\
         & \small (Accuracy \%) & \small (Accuracy \%) & \small (Win Rate \%) \\
        \midrule
        L-3-8B-Ins & 56.77 & \textbf{36.71} & \textbf{50.25} \\
        L\_CL\_Ins  & \textbf{57.01} & 36.25 & 49.75 \\
        \bottomrule
    \end{tabular}
    \caption{Model Performance Comparison on MMLU, LogiQA, and Alpaca\_Eval. Best scores are in \textbf{bold}.}
    \label{tab:english_performance}
\end{table}

\section{Case Study}
\label{sec:Error Case Studies}

To complement quantitative evaluations, we analyzed model outputs to identify performance patterns. Our findings align with LLM-assisted evaluations, confirming Llama\_CL's consistent superiority.

In MT, Llama-3-8B and Llama\_NTP occasionally produced source-language outputs---an issue largely mitigated in Llama\_Bi\_NTP and Llama\_CL. Llama\_CL showed fewer omissions and redundancies than Llama\_Bi\_NTP, indicating stronger cross-lingual comprehension. However, all models struggled with rare words and domain-specific terms in low-resource languages, revealing linguistic coverage limitations. For cross-lingual summarization, Llama\_CL sometimes defaulted to extractive summaries rather than abstractive content, exposing gaps in information synthesis. Sporadic Slovak usage appeared in Llama\_NTP, Llama\_Bi\_NTP, and Llama\_CL summaries, likely from monolingual corpus noise during continued pre-training. Despite improvements in cross-lingual QA, all models struggled with complex logical reasoning, highlighting persistent challenges in multilingual reasoning capabilities. Case study examples are in Tables~\ref{table:Translation Results Comparison},~\ref{table:CLSum Results Comparison}, and~\ref{table:CLQA Results Comparison}.

\begin{CJK*}{UTF8}{gbsn} 

\begin{table}[t!]
    \centering
    \footnotesize
    \renewcommand{\arraystretch}{1.6} 
    \setlength{\tabcolsep}{6pt} 
    \begin{tabular}{l p{2.2cm} p{2.2cm} p{1.4cm} p{2.2cm} p{1.4cm} p{2.2cm}}
        \toprule
        \textbf{Task} & \textbf{Question} & \textbf{L-3-8B} & \textbf{L\_NTP} & \textbf{L\_Bi\_NTP} & \textbf{L\_CL} & \textbf{Remark}\\
        \midrule
        \multirow{2}{*}{\centering MT} & Translate the following English to Chinese: "It also arguably makes reading easier, though writing is somewhat complicated by the need to find out whether a verb or adjective is used in a substantivized form." & It also arguably makes reading easier, though writing is somewhat complicated by the need to find out whether a verb or adjective is used in a substantivized form. & It also arguably makes reading easier, though writing is somewhat complicated by the need to find out whether a verb or adjective is used in a substantivized form. & 它也使阅读更容易，但由于需要确定动词或形容词是否使用了名词化形式，因此写作有些复杂。 & 它也可以说使阅读更容易，虽然写作是有点复杂的，因为需要找到一个动词或形容词是否在名词形式中使用。& Llama-3-8B and Llama\_NTP failed to translate the source sentence. \\
        \cmidrule(lr){2-7}
        & Translate the following English to Chinese: "According to the lawsuit, waste from the UN camp was not properly sanitized, causing bacteria to enter the tributary of the Artibonite River, one of Haiti's largest." & 根据诉讼,来自联合国营地的废物未得到适当消毒,导致细菌进入艺术博尼特河,海地最大的河流之一。 & 根据诉讼,联合国营地的垃圾没有得到适当的消毒,使细菌进入阿尔蒂波内特河,该河是海地最大的河流之一。 & 根据诉讼,联合国营地的垃圾没有得到适当的消毒,使细菌进入阿尔蒂波尼特河,是海地最大的河流之一。 & 根据诉讼，联合国营地的废物没有得到适当的消毒，导致细菌进入阿尔蒂博尼特河的分流之一，即海地最大的分流之一。& Llama\_CL generates a more detailed, accurate translation. \\
        \bottomrule
    \end{tabular}
    \caption{Translation Results Comparison}
    \label{table:Translation Results Comparison}
\end{table}

\end{CJK*} 

\begin{CJK*}{UTF8}{gbsn} 
\begin{table}[t!]
    \centering
    \footnotesize
    \renewcommand{\arraystretch}{1.6} 
    \setlength{\tabcolsep}{4pt} 
    \begin{tabular}{p{1cm} p{6cm} p{1cm} p{1cm} p{2cm} p{1cm} p{1cm}}
        \toprule
        \textbf{\centering Task} & \textbf{\centering Question} & \textbf{\centering L-3-8B} & \textbf{\centering L\_NTP} & \textbf{\centering L\_Bi\_NTP} & \textbf{\centering L\_CL} & \textbf{\centering Remark}\\
        \midrule
        \multirow{2}{*}{CLSum} & Please read the following English text and generate a short and precise Chinese summary: "By Peter BilesBBC World Affairs Correspondent The then-prime minister only saw it was likely after getting "raw intelligence" two days before the Argentines landed. Papers released under the 30-year rule show Mrs Thatcher was acutely worried about retaking the islands. One historian said the documents were among the "most powerful material" declassified in the last three decades. In October 1982, a few months after the war ended, Mrs Thatcher gave evidence behind closed doors to the Falkland Islands Review Committee, chaired by Lord Franks. The transcript of that dramatic testimony has now been published for the first time. "I never, never expected the Argentines to invade the Falklands head-on. It was such a stupid thing to do, as events happened, such a stupid thing even to contemplate doing", Mrs Thatcher told the Franks Committee. There had been some British contingency planning in the month before the Argentine invasion of the Falklands. On 26 March 1982, Ministry of Defence officials came back to Mrs Thatcher with a plan to deter a full-scale invasion. One sentence shocked her, and she wrote it in her diary: "Moreover, if faced with Argentine occupation on arrival there would be no certainty that such a force would be able to retake the dependency." She told the committee: "You can imagine that turned a knife in my heart, that lot." However, in her oral evidence, she said she had still considered an invasion unlikely: "I again stress, I thought that they would be so absurd and ridiculous to invade the Falklands that I did not think it would happen. "Nevertheless, one has always to make contingency plans, and soon after we got the South Georgia incident happening." The picture changed on 31 March 1982 when Mrs Thatcher was shown intelligence suggesting that an invasion was on the cards: "I just say it was the worst I think moment of my life." She also told the committee: "That night no-one could tell me whether we could retake the Falklands - no-one. We did not know - we did not know." The British foreign secretary at the time, Lord Carrington, also gave evidence to the Franks Committee, where he too held the view that Argentina was not going to invade the Falklands. "I have been accused, and was accused violently after 2 April in the House of Commons, and in the press and elsewhere - of wilfully ignoring signs and statements and evidence, and just ignoring it. I can truthfully say I did not do any of these things… they (the signs of an invasion) were not there," ... It also bases its claim on the islands' proximity to the South American mainland. Some 255 British servicemen and three Falklands civilians died during the conflict. The number of Argentine dead is estimated at about 650 All document images courtesy of the National Archives. & 玛格丽特·撒切尔前英国首相在1982年福克兰群岛战争之前就有所预料，但当时的政府官员却没有意识到这一点。 & 英国前首相玛格丽特·撒切尔说，她1982年对英国失去福克兰群岛的战争时期所面临的压力是"人生中最糟糕的一刻"。& 1982年4月2日，阿根廷突然入侵福克兰群岛，英国迅速集结军队回应。30年后，英国政府公布了关于这场战争的新档案。 & 马格丽特·撒切尔夫人当时说，她的生命从来没有经历过如此严重的危机。现在，刚公布的政府文件显示，这个英国首相在1982年阿根廷对福克兰群岛的入侵之前，并没有料到会发生这样的事。 & L\_CL lists two facts but does not generate a summary statement. 
        \\
        \bottomrule
    \end{tabular}
    \caption{CLSum Results Comparison}
    \label{table:CLSum Results Comparison}
\end{table}
\end{CJK*} 

\begin{table}[t!]
    \centering
    \footnotesize
    \renewcommand{\arraystretch}{1.6} 
    \setlength{\tabcolsep}{5pt} 
    \begin{tabular}{p{1cm} p{1.9cm} p{1.9cm} p{1.9cm} p{2.2cm} p{1.9cm} p{1.9cm}}
        \toprule
        \textbf{Task} & \textbf{Question} & \textbf{L-3-8B} & \textbf{L\_NTP} & \textbf{L\_Bi\_NTP} & \textbf{L\_CL} & \textbf{Remark} \\
        \midrule
        \multirow{2}{*}{\centering CLQA} & Answer the following questions in Czech. "Write a limerick about the joys of childhood in the style of Lewis Carroll." & Kdybys si vzpomněl na dětství, považte si to rádi!Bylo to tak plné nadšené vzrušení, plné úžasu a radosti!Nakonec to skončilo, ale jaká úžasná doba, když jsme byli malí a takoví, jaké jsme dnes. & Jako malý jsem si užíval,Až se mi v životě zatajvalo,Vždycky jsem věděl,Co jsem chtěl,Jako malé dítě, bylo to nejlepší.  & Jako malý holky v lese, \textbackslash nKterý nás potěšil,\textbackslash nTento svět byl plný štěstí,\textbackslash nS radostí a veselím, \textbackslash nNikdy jsme se necítily starosti. Overpostingotázkou& Koukal jsem na svět se štěstím plný,\textbackslash nA všude kolem se tají radost,\textbackslash nVysoké stromy, malé holky, \textbackslash nNávštěvy kamarádů, hrací kosti,\textbackslash nVšechno bylo úžasné, ne? To je pravda! & While both L\_CL and L\_Bi\_NTP employ parallel structure, a technique common in poetry, L\_CL's writing style is more poetic, with a rhythm that better aligns with the musical quality of verse. \\
        \cmidrule(lr){2-7}
        & Answer the following questions in Czech. "Explain to me the importance of the following quotation.'The only way to do great work is to love what you do.'" & Tento citát znamená, že pokud chcete dosáhnout úspěchu v jakékoli oblasti, musíte milovat to, co děláte. Kreativita a odhodlání jsou důležitými aspekty úspěchu, ale pokud se nebudete zajímat o to, co děláte, pravděpodobně nebudete schopni dosáhnout svých cílů. Mít opravdový zájem o své pracovní povinnosti je klíčem k tomu, abyste se dostali k tomu, kde se chcete dostat. & To love what you do je důležitý aspekt úspěšné práce. Když si člověk vybere práci, kterou opravdu miluje, je mnohem pravděpodobnější, že se mu podaří vykonávat ji s velkým nasazením a přinášet kvalitní výsledky. Milovat svou práci zajišťuje, že člověk se bude cítit spokojenějším a bude mít větší motivaci a nadšení k práci. Navíc, když se člověk skutečně zajímá o svou práci, je pravděpodobnější, že bude mít větší znalosti a schopnosti a bude lépe splňovat požadavky a očekávání zaměstnavatele. & To mít velký úspěch ve svém povolání, je třeba milovat to, co děláte. Když máte rádi to, co děláte, máte motivaci pracovat tvrdě, vyvíjet se a dosahovat cílů. Milujete své povolání, budete se cítit spokojenější a vaše práce se bude lépe projevovat v práci a životě. & To nejlepší, co můžete dělat, je milovat to, co děláte. Pokud vás vaše práce nudí, vaše výsledky budou značně omezené. Když milujete svou práci, získáte z ní větší spokojenost a z ní se vám líbí být. To vám umožní dělat svou práci s větší odvahou a odhodláním a získáte více z jejího výkonu. & Responses of L\_CL and L-3-8B are relatively accurate, with the answer of L\_CL answer being more relevant to the question. The response of L\_NTP contains English content, while the answer of L\_Bi\_NTP has repetitive elements.  \\
        \bottomrule
    \end{tabular}
    \caption{CLQA Results Comparison}
    \label{table:CLQA Results Comparison}
\end{table}

\section{Conclusion}

We introduced a novel continued pre-training approach that enhances cross-lingual capabilities in multilingual LLMs by integrating cross-lingual mapping tasks alongside language-specific next-token prediction. This dual-task strategy improved language alignment while preserving monolingual fluency, addressing key limitations in existing multilingual models.

Extensive experiments validated our approach, demonstrating improvements in language alignment, monolingual perplexity, and cross-lingual task performance. Our proposed LAC offers a robust metric for assessing inter-language coherence, even with limited test data. Additionally, our cross-lingual open-domain QA dataset and LLM-assisted evaluation highlight the effectiveness of explicit cross-lingual pre-training in enhancing generative capabilities across seen and unseen language pairs.

While our method significantly improves cross-lingual performance, challenges remain in summarization and complex reasoning tasks. Future work should explore better integration of summarization objectives and enhanced multilingual reasoning. Expanding our framework to underrepresented languages will further strengthen multilingual LLMs, reinforcing the value of explicitly modeling cross-lingual relationships in pre-training.

\section{Limitation}

While our approach significantly improves the cross-lingual generation and comprehension capabilities of multilingual LLMs, certain limitations remain. Notably, we observed only marginal improvements in tasks requiring cross-lingual text summarization and complex reasoning. This indicates that while our method enhances language alignment and generation fluency, it struggles with tasks that demand deeper information extraction, filtering, and logical reasoning across languages. These limitations may arise from the inherent complexity of summarization and reasoning tasks, which require more sophisticated mechanisms to retain and manipulate detailed semantic information across languages.

Additionally, although our approach effectively handles low-resource languages and demonstrates strong alignment for bilingual tasks, the gains in highly subjective or culturally nuanced tasks, such as humor or sentiment analysis, are less pronounced. Addressing these challenges will be a focus of future research, where we aim to develop methods that strengthen cross-lingual logical reasoning and enhance the model's capacity for processing complex, abstract concepts.

\begin{acks}
This research is supported by the National Research Foundation, Singapore under its National Large Language Models Funding Initiative. Any opinions, findings, conclusions, or recommendations expressed in this material are those of the author(s) and do not reflect the views of the National Research Foundation, Singapore. The research is also supported by the National Research Foundation, Singapore under its National Large Language Models Funding Initiative (AISG Award No: AISG-NMLP-2024-005 and AISG-NMLP-2024-004).
\end{acks}

\bibliographystyle{ACM-Reference-Format}
\bibliography{sample-base}

\end{document}